\documentclass[journal]{IEEEtran}
\usepackage{packages}
\usepackage{color}
\title{TMIXT: A process flow for Transcribing MIXed handwritten and machine-printed Text }

\author[1]{Fady Medhat}
\author[1]{Mahnaz Mohammadi}
\author[1]{Sardar Jaf}
\author[1]{Chris G. Willcocks}
\author[1]{Toby P. Breckon}
\author[1]{Peter Matthews}
\author[2]{Andrew Stephen McGough}
\author[3]{Georgios Theodoropoulos}
\author[(\Letter)1]{Boguslaw Obara}

\affil[1]{\footnotesize Department of Computer Science, Durham University, UK}
\affil[2]{\footnotesize School of Computing, Newcastle University, UK}
\affil[3]{\footnotesize Department of Computer Science and Engineering, Southern University of Science and Technology, China}

\vspace{50cm}
\affil[1]{\textit {\{fady.medhat, mahnaz.mohammadi, sardar.jaf, christopher.g.willcocks, toby.breckon, p.c.matthews, boguslaw.obara\}@durham.ac.uk}}
\affil[ ]{\textsuperscript{2}\textit{stephen.mcgough@newcastle.ac.uk}\hspace{0.2cm} \textsuperscript{3}\textit {georgios@sustc.edu.cn}}

\begin{document}
\maketitle

\providecommand{\keywords}[1]
{
  \small	
  \textbf{\textit{Keywords---}} #1
}

\begin{abstract}
Handling large corpuses of documents is of significant importance in many fields, no more so than in the areas of crime investigation and defence, where an organisation may be presented with a large volume of scanned documents which need to be processed in a finite time. 
However, this problem is exacerbated both by the volume, in terms of scanned documents and the complexity of the pages, which need to be processed. Often containing many different elements, which each need to be processed and understood. 
Text recognition, which is a primary task of this process, is usually dependent upon the type of text, being either handwritten or machine-printed.
Accordingly, the recognition involves prior classification of the text category, before deciding on the recognition method to be applied. 
This poses a more challenging task if a document contains both handwritten and machine-printed text. 
In this work, we present a generic process flow for text recognition in scanned documents containing mixed handwritten and machine-printed 
text without the need to classify text in advance. We realize the proposed process flow using several open-source image processing and text recognition packages\footnote{Code: https://github.com/fadymedhat/TMIXT}.
The evaluation is performed using a specially developed variant, presented in this work, of the IAM  handwriting database, where we achieve an average transcription accuracy of nearly 80\% for pages containing both printed and handwritten text.

\vspace{4mm}
\end{abstract}

%------------------------------------------------------------------------------%
\section{Introduction}\label{Introduction}
\noindent Despite the migration to fully electronic administration system across civil and non-civil sectors, most businesses and governmental agencies hold a significant quantity of historical archive material. Originally these documents would have been in the form of printed documents\footnote{We do not distinguish here between different printing processes or even typed text as these are handled through the same process.}, handwritten forms, scanned pages, etc. The availability of such documents is not only confined to closed entities like businesses or agencies, but it also extends to documents proliferating through cyberspace. For example, organizations such as Wikileaks reports that it has collected over 10 million documents, since 2006, involving war, spying and corruption. The sheer volume of documents imposes a need to automate the transcription process to machine readable formats for further security analysis and defence related applications.   
Since such information is largely made up of text, Natural Language Processing (NLP) for big data presents an opportunity to take advantage of what is contained in these documents and also reveal patterns, connections and trends across disparate sources of data.
NLP techniques incorporate a variety of methods, including linguistics, semantics, statistics and machine learning to extract entities, relationships, and  context, which enables an understanding of what is being written, in a comprehensive way. Far beyond what could be achieved through analysis by individuals.

A database of scanned text documents is one of the major examples of data sources, where Optical Character Recognition (OCR) systems, being concerned with the semantics recognition of the NLP problem, are used to extract the contents of such documents in order to convert them into a machine readable format to facilitate the retrieval and re-usability of the knowledge held within. These documents may include either handwritten, machine-printed text or both.
The respective recognition of handwritten and machine-printed documents are commonly handled separately using dedicated techniques due to the inconsistency in the structure of characters and the style of handwritten text when compared to machine-printed text.

Text localization followed by classification (into hand-written or machine printed) are two important stages before the text recognition phase. Several attempts \cite{guo2001separating, zheng2004machine, kandan2007robust, chanda2010structural, belaid2013handwritten} that considered these stages depend mostly on 
hand-crafted features and most of these approaches are used in combination with conventional machine learning classifiers such as k-Nearest Neighbor (k-NN) and Support Vector Machines (SVM) \cite{belaid2013handwritten}. Other attempts exploited the use of statistical models. For example, Cao {\em et al.}~\cite{cao2011handwritten} 
presented a system for identification and classification of handwritten and machine-printed\footnote{Sometimes referred to as typewritten, we will use machine-printed here without loss of generality.} text from document images using Hidden Markov Models (HMM). Silva {\em et al.}~\cite{da2009automatic}, proposed an automatic discrimination system for identifying handwritten words from machine-printed words. 
They employed several preprocessing steps for image segmentation, feature extraction and finally classification. The effort of Zagoris {\em et al.}~\cite{zagoris2014distinction}, despite being based on a traditional flow of image preprocessing and feature extraction, differs in being dependent on a Bag of Visual Words (BoVW) model.
This model depends on creating a codebook for features extracted from a training dataset that can be further used to generate a code for a new text block. 
The vectors generated using the code book are further classified using a combination of binary SVM classifiers to decide between handwritten, machine-printed and noise. 

Classification methods being rule-based, structural or statistical \cite{plamondon2000online} have been used in combination with hand-engineered features, where HMM were used to capture the feature transition of the handwritten text \cite{el1999hmm}. 
Other attempts have tried to combine both HMM with neural networks as in \cite{krevat2006improving}. Later attempts tried to exploit the use of neural networks as feature extractors especially using Recurrent Neural Networks (RNN) and their more advanced counterpart, the Long Short-Term Memory (LSTM). Graves {\em et al.}~\cite{graves2009offline,graves2009novel}
extended the  use of a single dimensional LSTM to a multidimensional one, where they achieved competitive performance compared to HMM.

Convolutional Neural Networks (CNN) \cite{lecun1998gradient} have been used in many text classification tasks. Feng {\em et al.}~\cite{feng2017robust} have addressed the problem of machine-printed and handwritten text separation using a CNN model. The text-line input to their CNN model captured discriminative content for the classification task. Their experimental validation also showed that integrating their proposed cropping schemes with deep architectures and wider convolutional filters improved the performance significantly.

Most of the referenced attempts are targeted for classifying text as either handwritten or machine-printed text, and the recognition process is applied with the prior knowledge of the type of text under consideration. This imposes an identification overhead especially when the number of documents is large. The presence of both types of text in a document introduces even more challenges for OCR systems, since it requires reliant localization and segmentation methods for different regions within a page upon which a classifier will be able to distinguish between machine-printed and handwritten text for later recognition. 

In contrast to most prior works in this field, we propose a generic process flow for Transcribing scanned documents containing MIXed handwritten and machine-printed Text (TMIXT) without the need for any prior knowledge of the type of text within the document, which is the key novelty of this work. TMIXT is implemented using several publicly available packages to fulfill the designated task of each stage within the process flow. The proposed process flow is evaluated using the IAM handwriting database\cite{marti2002iam}, widely used for text recognition research, in combination with a set of specially tailored labels, the MIXED-IAM\footnote{\url{https://bitbucket.org/DBIL/mixed-iamdb}} labels we present in this work, to cope with the nature of mixed text documents.

The rest of this paper is organized as follows. Our proposed process flow for scanned text recognition and its constituent modules are discussed in detail in Section \ref{Methodology}. We present an analysis of different preprocessing stages, and evaluate the performance of our proposed process on the IAM database using different evaluation metrics in Section \ref{Experiments}, and conclude the paper and discuss future works in Section \ref{Conclusions and Future Work}.
%------------------------------------------------------------------------------%
% \input{tex/literature}
%------------------------------------------------------------------------------%
%\vskip -0.15in
\begin{figure}
\vskip 0in
\begin{center}
\centerline{\includegraphics[width=8cm]{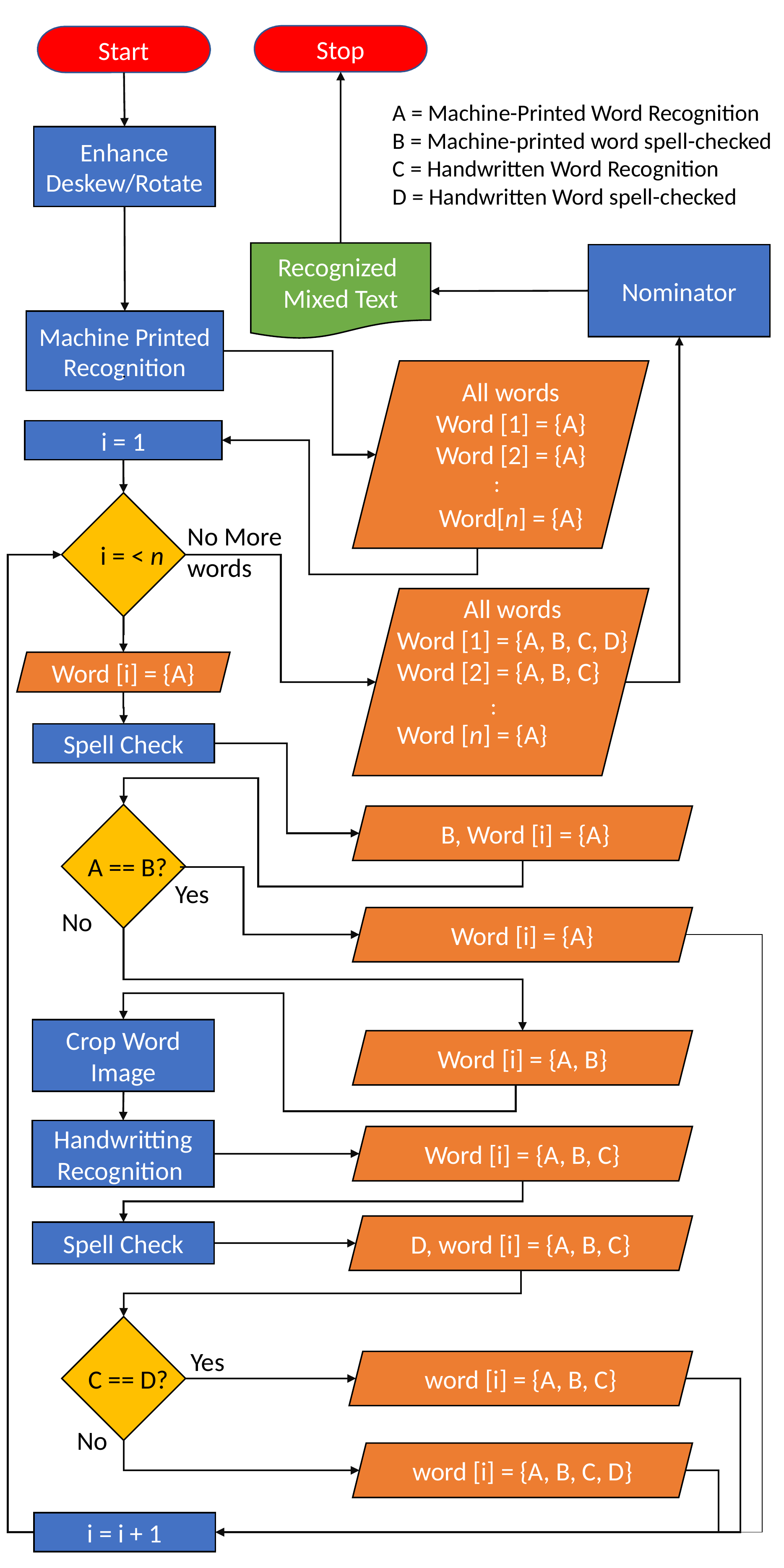}}
%\vskip -0.1in
\caption{TMIXT process flow for scanned document recognition.}
\label{fig:processflow}
\end{center}
\vskip -0.4in
\end{figure} 
\section{Methodology}\label{Methodology}
\noindent The TMIXT process flow, we present in this work, allows for transcribing mixed handwritten and machine-printed text without the need to discriminate the handwritten and machine-printed text prior to recognition.  
This process flow is realized through a suite of open-source software components that can be interchanged with other libraries and packages of similar functionality within the overall architecture of our methodology. 

We will discuss the general flow of our process flow before presenting a more detailed discussion of each sub-part in the relevant subsection below. Following Fig. \ref{fig:processflow}, a single scanned image of a text document undergoes several phases in the TMIXT process flow. The first phase involves simple image processing and enhancement, that could have direct influence on the recognition performance. The enhanced page is forwarded to the machine-printed text recognition. This stage is applied on the whole page irrespective of the actual text type(s), being handwritten or machine printed, present in the page. Spell-checking is applied on each word generated, where failure to pass the spell-checking validation induces the next phase of the process flow to be performed -- the handwriting recognition.

The handwriting recognition is applied at word-level compared to the page-level used in the machine-printed recognition. This is fulfilled by cropping the exact word that fails the spell-checking, and applying the handwriting recognition on this specific word in isolation from the rest of the page. Accordingly, each word in the document could have up to four possible options generated from the machine-printed and the handwriting recognition phases together with their spell-checking. These options are subject to an elimination process to nominate the optimum candidate word based on the context, which is the role of nomination phase -- the final phase. Since this work is focusing on presenting the TMIXT flow without constraining the libraries used in implementation, we will avoid details related to training the specific recognition models or the internals of the image enhancement algorithms we adopted for this work. The following subsections discuss each stage of the process flow in more detail.

\subsection{Preprocessing}
\noindent The preprocessing stage aims to eliminate variations such as deformations and noise in scanned images, which affect the recognition accuracy.

\subsubsection{Image Enhancement}

\begin{figure}[t!]
\vskip 0in
\begin{center}
\centerline{\includegraphics[width=9.5cm]{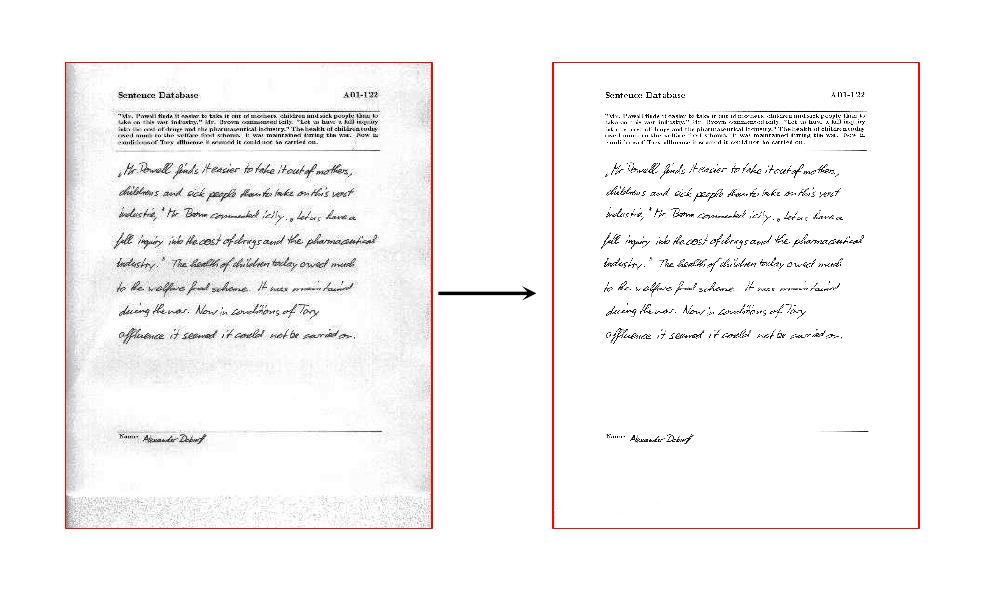}}
\vskip -0.1in
\caption{Enhancement of a Scanned Document.}
\label{fig:enhancedpages}
\end{center}
\vskip -0.3in
\end{figure}

Enhancing an image is the first stage of the proposed flow as shown in Fig. \ref{fig:processflow}. This stage involves noise and artefacts removal in addition to increasing the contrast between the text and the background. A survey by Jung {\em et al.}~\cite{jung2004text} investigated a wide range of proposed approaches for image enhancement to eliminate distortions in images and documents. Filter-based methods are widely used, which are mostly dependent on classical image processing techniques. Fig. \ref{fig:enhancedpages} shows a sample page before and after enhancement.

\subsubsection{Angular Alignment}
A page could be scanned with random rotation degrees, which introduces a form of skewness in the scanned image. Loading a skewed image directly to a recognition system degrades the recognition accuracy. A range of attempts have been proposed to tackle this problem \cite{al2015review}. For example, in Profile Projection (PP) analysis, the image is projected to a single vector and further analysis is applied to estimate the skewness angle. The Hough Transform \cite{osti_4746348} is one of the most widely adopted methods for skew detection and has been especially used for text lines \cite{likforman1995hough,louloudis2006block}.

\begin{figure}[t!]
\vskip 0in
\begin{center}
\centerline{\includegraphics[width=9.5cm]{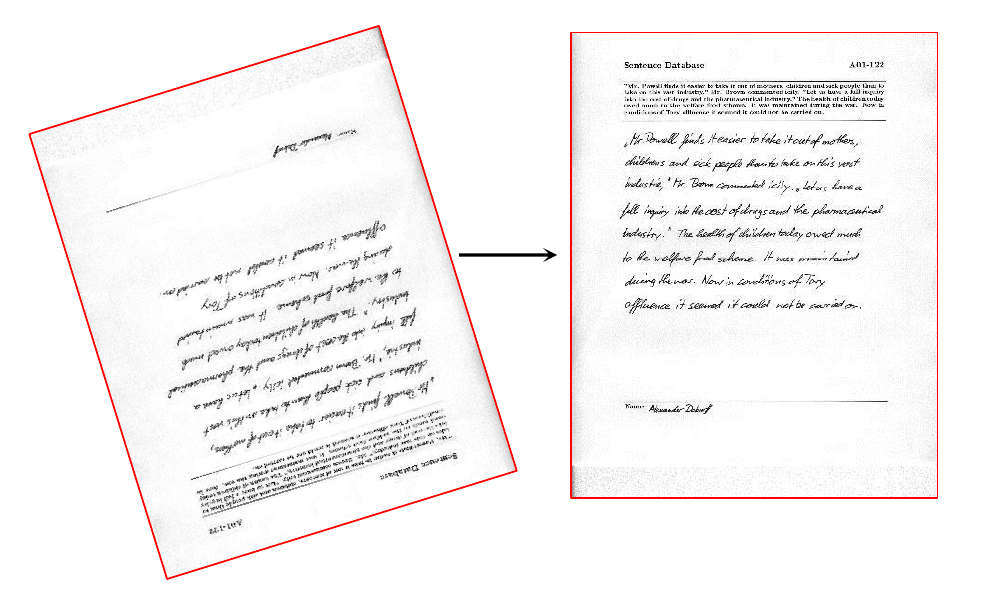}}
\vskip -0.1in
\caption{Angular Alignment of a Scanned Document.}
\label{fig:deskewedpages}
\end{center}
\vskip -0.3in
\end{figure}

In our proposed architecture, the angular alignment follows a two-step approach: first we deskew a page to convert a near-vertical (e.g. 13$^{\circ}$ inclination) or a near-horizontal page to be either strictly vertical or horizontal. We also assume that a page could be rotated by a cardinal angle. Accordingly, the second step involves rotating the image with 0$^{\circ}$, 90$^{\circ}$, 180$^{\circ}$ and 270$^{\circ}$ degrees. For each of the rotated variants of an image, regardless of the content type (handwritten or machine-printed), of the scanned document we apply a machine printed text transcription and score the generated words for each of the four rotations against a dictionary. The rotation with the highest score is assumed to be the optimum rotation of the page, which is further considered for the rest of the processing stages. Fig. \ref{fig:deskewedpages} shows the angular alignment applied on an input image. We assume here that the page has one dominant rotation for text. However, if this is not the case the page could be divided into regions which are processed independently for rotation.

\subsection{Machine-Printed Text Recognition}

\noindent Text recognition has been studied widely \cite{impedovo1991optical,ye2015text}. The work of Smith \cite{smith1987extraction} proposed a complete framework for text recognition that involved several methods for text detection, localization and recognition. His work became the basis of one of the most successful text recognition frameworks, Tesseract~\cite{smith2007overview}, accordingly we used it in our work.

\begin{figure*}
\vskip 0in
\begin{center}
\centerline{\includegraphics[width=\textwidth]{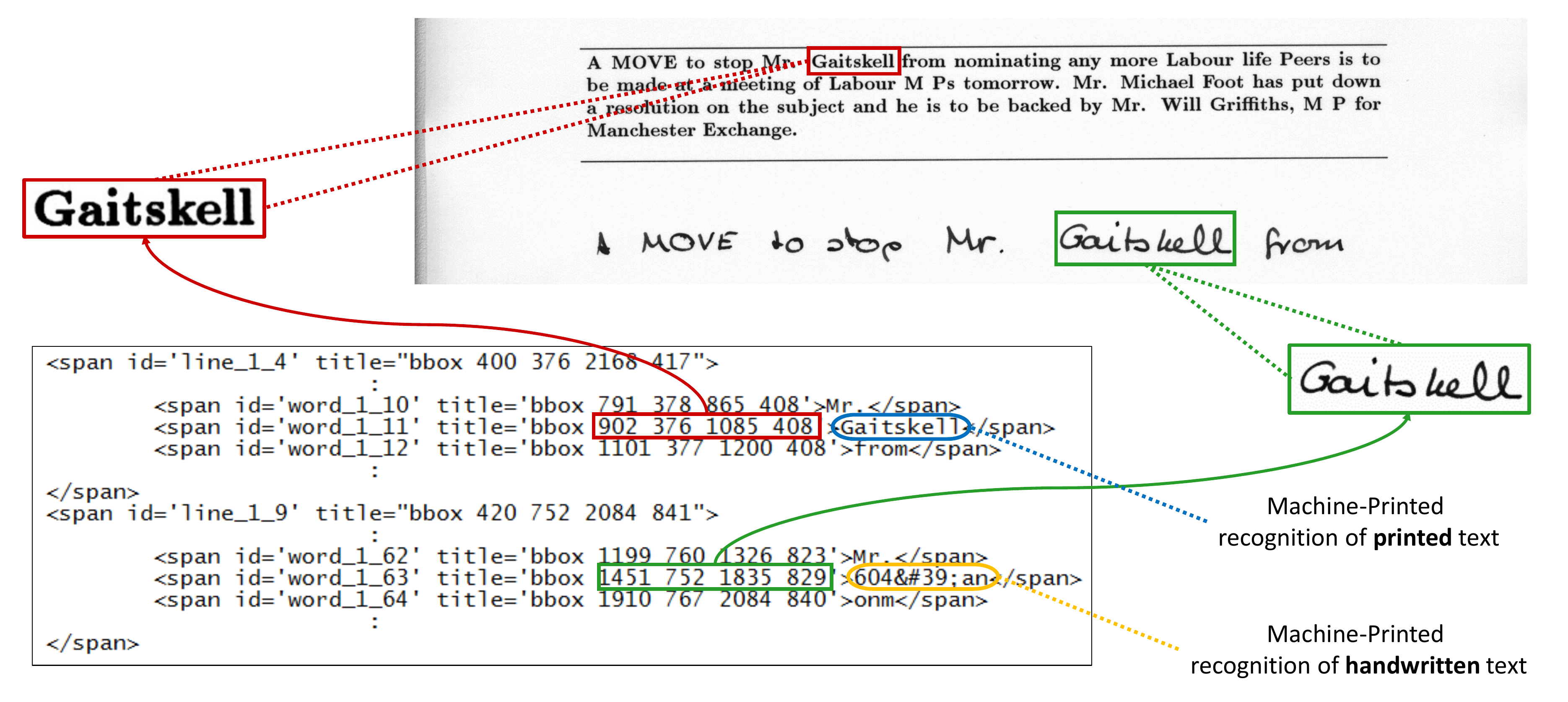}}
\vskip -0.1in
\caption{An hOCR File Containing Recognized Text and The Bounding-box for Each Word.}
\label{fig:pagecropping}
\end{center}
\vskip -0.1in
\end{figure*}

Despite the high accuracy of Tesseract in transcribing the machine-printed text, it fails in transcribing the handwritten one. 
For example, Fig. \ref{fig:pagecropping} shows a machine-printed word, ``Gaitskell'', correctly recognized while its corresponding
handwritten recognition is ``604\&\#39;an''. In our process flow, we used the low level transcription accuracy, detected by matching the generated word against the spell-checked version, as an indication of the presence of handwritten text in the input image.
\subsection{Spell Checking}
\noindent The problem of spell checking and word correction has been considered for several decades \cite{kukich1992techniques}. Hodge {\em et al.}~\cite{hodge2003comparison} tried to merge the performance of phonetic and associative matching in addition to supervised learning methods to develop a hybrid method for spelling correction. Simple correction methods involve constructing a tree of a language vocabulary with the help of a predefined dictionary, and they proceed by finding matches of the most similar words to the word being corrected. Other methods exploit the use of a language model and the word context \cite{carlson2007memory} to enhance the correction decision. 

The spell-checking stage is used in two phases within the process flow. Firstly it is used after the machine-printed text recognition and based on its output, the handwritten text recognition is induced. Secondly, it is also used after the handwritten text recognition stage to validate the handwritten text recognition output. We used a combination of two spell-checkers \cite{autocorrect, enchant}, more could be included to improve the correction decision. 

\subsection{Word Cropping}

\noindent In the cropping stage, words that fail the spell-checking test of the machine-printed text recognition phase are extracted using the bounding box defined for each word. We found through empirical trials that the recognition accuracy is degraded when the boundary characters reside over the border of the cropped image. This is due to the inability of the recognition model to capture the actual vertical and horizontal context of the pixels at boundaries of a word. To address this issue, we padded the cropped word with white space in all directions.

\subsection{Handwritten Text Recognition}

\noindent Statistical methods such as HMM have been used for handwritten text recognition. The HMM was also considered in combination with neural networks in \cite{espana2011improving}. A two dimensional HMM to capture the changes in both the horizontal and vertical axes of a text line was investigated in \cite{park1998truly}. Neural network based models have also been applied to the problem. Long Short-Term Memory (LSTM) \cite{hochreiter1997long} is one of the successful neural based models that is adapted for temporal sequences especially text, where Graves {\em et al.}~\cite{graves2009offline,graves2009novel} proposed a multidimensional LSTM for text recognition. Motivated by the success of RNN in text recognition, we adopted models dependant on LSTM for this stage.

\subsection{Word Nomination}

\noindent Following Fig. \ref{fig:processflow}, at the nomination stage we are aiming to retrieve the optimum candidate word from the list of options generated through the process flow for each single word present in the document. For this task, we propose a rule-based and a context-based method.

Each potential word in the document at this stage will have a maximum of four options (A, B, C, D), which are the machine-printed text recognition output (A) and its spell-checked version (B), and the handwritten text recognition version (C) together with its spell-checked output (D). 
The list of options could be of size one (if B matches A) or three (if D matches C) or four (if D does not match C).

\begin{algorithm}[!t]
\small
\caption{Rule-Based Nomination to Retrieve the Optimum Candidate Word}
\label{alg:rule}
\begin{algorithmic}[1]
\State $CO : Candiate~Option$
\State $Options~List : [A, B, C, D]$\\
\If{$Options~Count == 1$}
  \State $CO = A$
\ElsIf{$Options~Count == 3$}
       \State $CO = C$
\ElsIf{$Options~Count == 4$}
  \If{D \! != \! \textless UNK\textgreater}
       \State $CO = D$
    \Else   
      \If{B \! != \! \textless UNK\textgreater}
	\State $CO = B$
      \Else   
	\State $CO = A$
      \EndIf
    \EndIf
\EndIf
\end{algorithmic}
\end{algorithm}

\begin{figure*}[!t]
\vskip 0in
\begin{center}
\centerline{\includegraphics[width=\textwidth]{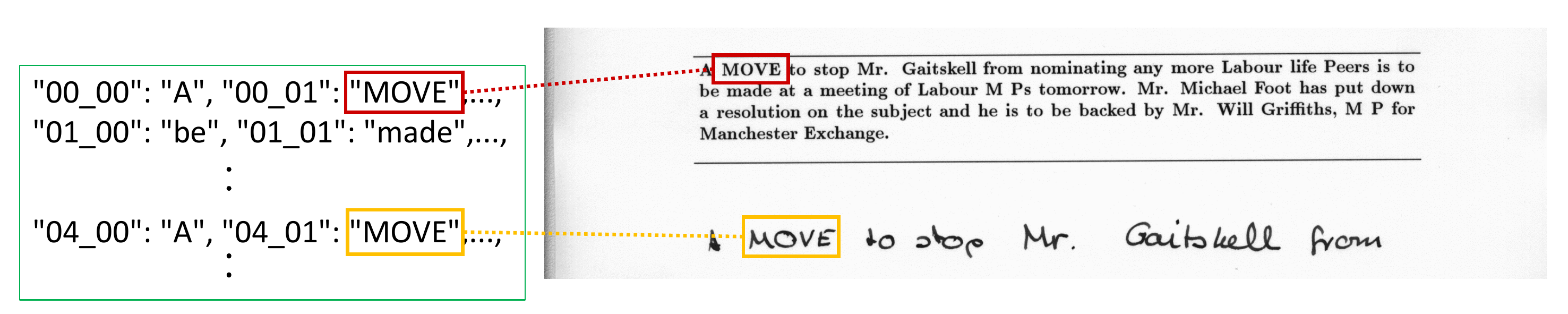}}
%\vskip -0.1in
\caption{The format of the MIXED-IAM labels for a sample page of the IAM database.}
\label{fig:mixediam}
\end{center}
\vskip -0.1in
\end{figure*}   

\begin{algorithm}%[!t]
\small
\caption{Context-Based Nomination to Retrieve the Optimum Candidate Word}
\label{alg:context}
\begin{algorithmic}[1]
\State $CO : Candiate\,Option$
\State $PoL: Previous\,options\,List$
\State $CoL: Current\,options\,List$
\State $NoL: Next\,options\,List$
\State $SM: Similarity\,Matrix$\\
\State $PCoL = CrossJoin (PoL,CoL)$
\State $CNoL = CrossJoin (CoL,NoL)$
\State $PCoEL = Embeddings (PCoL)$
\State $CNoEL = Embeddings (CNoL)$\\
\For{$i < len(PCoEL)$}
 \For{$j < len(CNoEL)$}
 \hspace{-5cm}\State $SM[i][j] = Cosine(PCoEL[i],CNoEL[j])$
 \EndFor
\EndFor\\
\State $index = argmax(argmax(SM))$
\State $CO = CoL[index]$
\end{algorithmic}
\end{algorithm}

In the rule-based nomination, Alg. \ref{alg:rule}, Option A is considered as the candidate option by default, and based on the list size it could be either substituted or chosen as the candidate option. For example, if the list has three options, it means that the machine-printed version has failed the spell-checking stage accordingly the system had to retrieve the handwritten variant of the word, which passed the spell-checking. In this case, the third option (C) for the handwritten version is the chosen one. A similar rule applies to a list of four options, but in this case we have to consider the possibility of having an unknown flag generated from the spell-checking of either the machine-printed and the handwritten variant as listed in Alg. \ref{alg:rule}.

In the context-based nomination, Alg. \ref{alg:context}, choosing from one of the possible four options of each word is considered within a context of one preceding and one succeeding word. The elimination stage would have been straight forward if the neighbouring words in a context do not have an option list, but the case could be that each of these words have an option list on its own, which introduces additional complexity to the decision process. To fulfill the nomination process in such scenario, bi-grams are constructed from each of the words in the current options list and all the words in the options list of the preceding and succeeding words. For example, for a current word options list of size $4$ and a succeeding word options list of size $3$ (noting that the preceeding options list will always contain a single word due to the progression order of the transcription), we will have a total of $4$ bi-grams between the current and the preceding words and $12$ bi-grams between the current and the succeeding words. Bi-gram embedding vectors are retrieved using pre-trained models base on \cite{mikolov2013efficient}. The distances between all combinations of vectors of the previous embeddings set and the succeeding embeddings set of vectors are retrieved. The smaller the distance, the more it indicates a higher probability of a pair of bi-grams to reside in proximity to each other in the feature space. The smallest distance is used in choosing the optimum candidate from the options list. Alg. \ref{alg:context} summarizes the operation of the context-based nomination, where a cross join operation is performed between the options list of a preceding word (PoL) and the options list of  the current word (CoL). A similar cross join is applied to current options list and the succeeding options list (NoL). Following the generation of the cross join, the embeddings are generated for each bi-gram item. The embeddings of the previous-current options (PCoEL in Alg. \ref{alg:context}) are compared against each item in the current-next (CNoEL in Alg. \ref{alg:context}) options using cosine similarity between each pair of vectors. The index of the vector with maximum similarity value is the index of the optimum candidate word in the current option list.
%%---------------------------------------------------------------------------- %

\section{Experiments}\label{Experiments}
\noindent In this section, we evaluate the accuracy of our proposed flow and provide analysis to the effect of different stages in the process flow.

%\begin{figure}[!t]
%\vskip -0.05in
% \hrulefill
% 
%     "00\_00": "A", "00\_01": "MOVE",...,\\ 
%     "01\_00": "be", "01\_01": "made",...,\\
%     "02\_00": "a", "02\_01": "resolution",...,\\
%     "03\_00": "Manchester", "03\_01": "Exchange",...,\\
%     "04\_00": "A", "04\_01": "MOVE",..., \\
%     "05\_00": "nominating", "05\_01": "any",...,\\
%     "06\_00": "is", "06\_01": "to",...,\\
%     "07\_00": "M Ps", "07\_01": "tomorrow",..., \\
%     "08\_00": "put", "08\_01": "down",..., \\
%     "09\_00": "and", "09\_01": "he",..., \\
%     "10\_00": "Griffiths",..., "10\_05": "Exchange"
%     
%\hrulefill
%\caption{The format of the MIXED-IAM labels for a sample page of the IAM database.}
%\label{fig:mixediam}
%\vskip -0.2in
%\end{figure}

\subsection{MIXED-IAM labels}

\noindent The IAM handwriting database \cite{marti2002iam} is composed of 1539 forms of handwritten English text written by 600 writers. 
A single page of the IAM database is split into two sections, machine-printed text in the upper section and the bottom section of the page is used for a writer to fill with their style of handwriting -- transcribing the machine-printed text above\footnote{Although we know what the handwritten text is supposed to say we do not use that information in our work.}. The dataset is released with tokenized transcription for each text line of the handwriting section. Paragraph-level transcription for the machine-printed section of the page is also provided along with the database.

We used Natural Language Processing Toolkit (NLTK) \cite{bird2004nltk} to tokenize the machine printed section and merged both sections into a single transcription represented by the lines and word order in the page disregarding the word type as shown in Fig. \ref{fig:mixediam}.  
 
\subsection{Evaluation Metrics}

\noindent The output transcription of the TMIXT process flow is concatenated into a single paragraph, and similarly the corresponding target transcription within the MIXED-IAM labels.
Both represent the predicted and target transcription for an image of a document retrieved from the IAM database, respectively.

The evaluation we applied is based on:
\begin{itemize}
 \item Character level evaluation using the Levenshtein distance \cite{levenshtein1966binary} to measure the number of deletions, insertions, or substitutions required to transform the predicted document to the target document.
  \item Word level evaluation using a Bag-of-Word (BoW) \cite{salton1975vector} representation, which involves comparing the frequency of occurrences of words in either the predicted or the target transcription disregarding the syntax structure.
  \item Document similarity, which involves generating an embedding vector \cite{pgj2017unsup} for the complete document for either the prediction or the target transcription, and using a similarity measure between the generated embeddings.  
\end{itemize}

\subsection{Libraries and packages}
\noindent We used several open-source efforts to implement the TMIXT process flow. For the image enhancement and skewness stage we used the work
in \cite{villegas2015modification} and \cite{deskew}, respectively. We used Pillow \cite{pillow} for the cropping and padding of images. We used Tesseract \cite{smith2007overview} for the machine-printed and Laia \cite{laia2016} for the handwriting text recognition. Tesseract was also used to generate the hOCR files containing the bounding boxes of the words to be cropped. We used the work of Pagliardini {\em et al.}~\cite{pgj2017unsup} for the embeddings generation of the bi-grams and for the spell-checking we used \cite{autocorrect} and \cite{enchant}.

%https://link.springer.com/content/pdf/10.1007%2F978-3-319-73165-0_8.pdf

%\textit%\begin{table}[!htb]
%\caption {Number of Documents in Different Accuracy Ranges Using Rule-Based Elimination Method}
%\label{tab:table4}    
%\centering
%\pgfplotstabletypeset[
%color cells={min=-0,max=1087},
%col sep=comma,
%columns/Accuracy Range/.style={reset styles,string type}
%]{
%       Accuracy Range,Levenshtein, String Similarity
%       90\%-100\% ,48,  24
%       80\%-90\%,758, 300
%       70\%-80\%, 592, 547
%       60\%-70\%,110, 326
%       50\%-60\%,24,  215
%       40\%-50\%, 7 ,  99 
%       30\%-40\%, 0 ,   26
%       20\%-30\% , 0 ,   2
%}
%\end{table}
%\section{Results}
%{\color{blue}moved from previous section as these are also results.}
\begin{figure}[t]
\begin{center}
\centerline{\includegraphics[width=9cm]{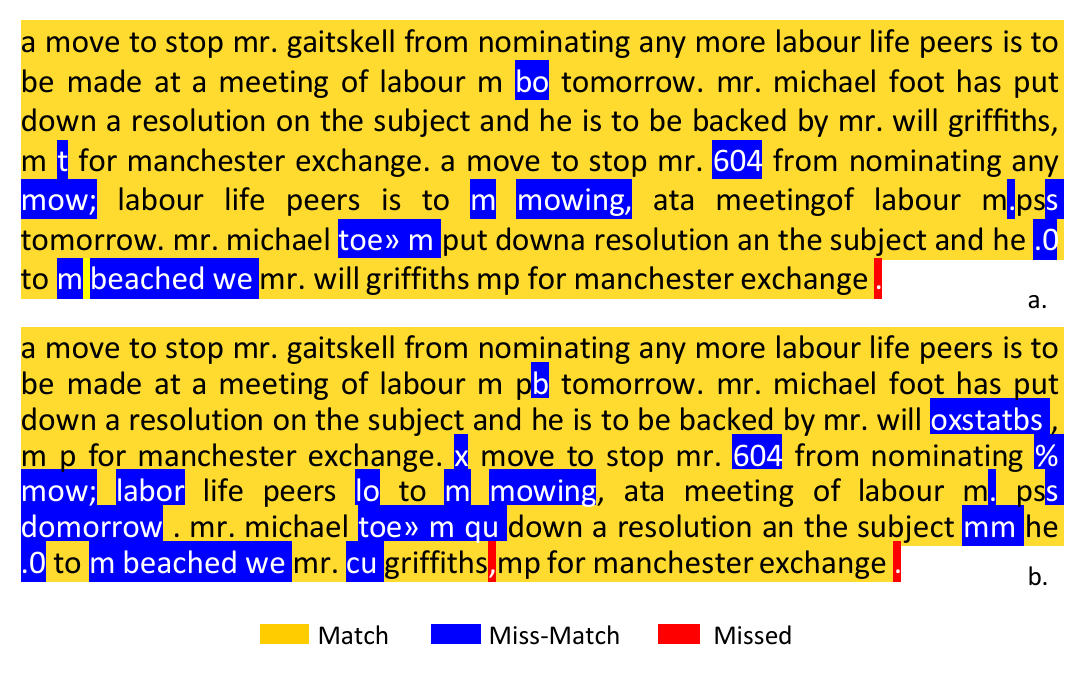}}
\vskip -0.1in
\caption{Effect of enhancement on text recognition: a. with enhancement, b. without enhancement.}
\label{fig:enhancement}
\end{center}
\vskip -0.3in
\end{figure} 
\begin{figure}[t]
\begin{center}
\centerline{\includegraphics[width=9cm]{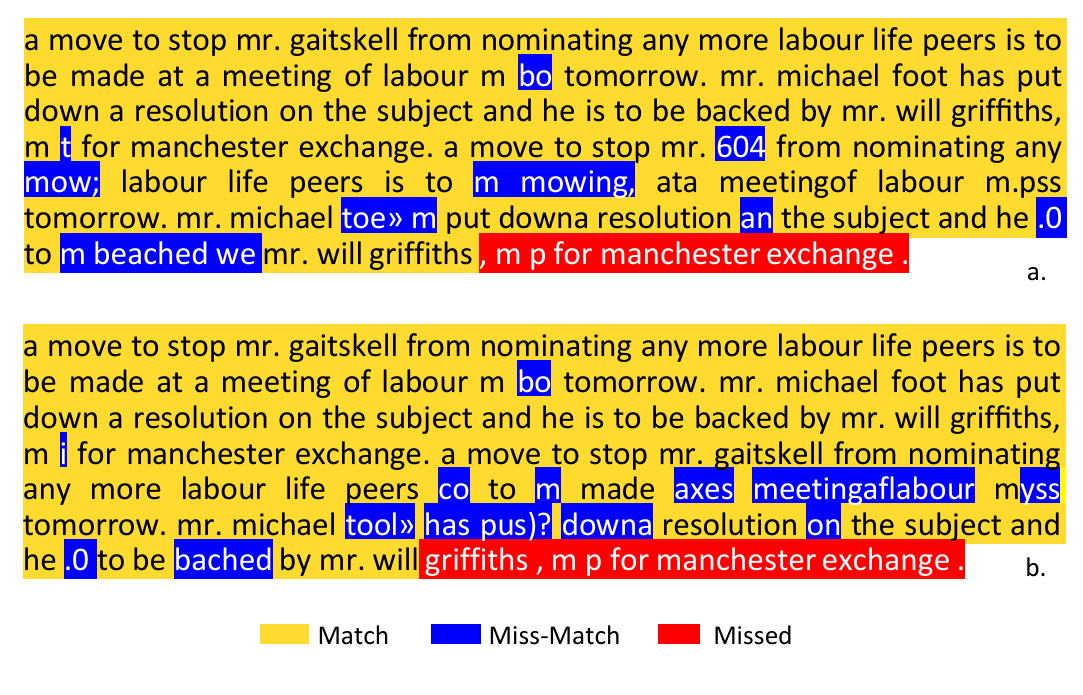}}
\vskip -0.1in
\caption{Effect of deskewness on text recognition: a. deskewed, b. without deskewness.}
\label{fig:skew}
\end{center}
\vskip -0.3in
\end{figure} 
\begin{figure}[t]
\begin{center}
\centerline{\includegraphics[width=9cm]{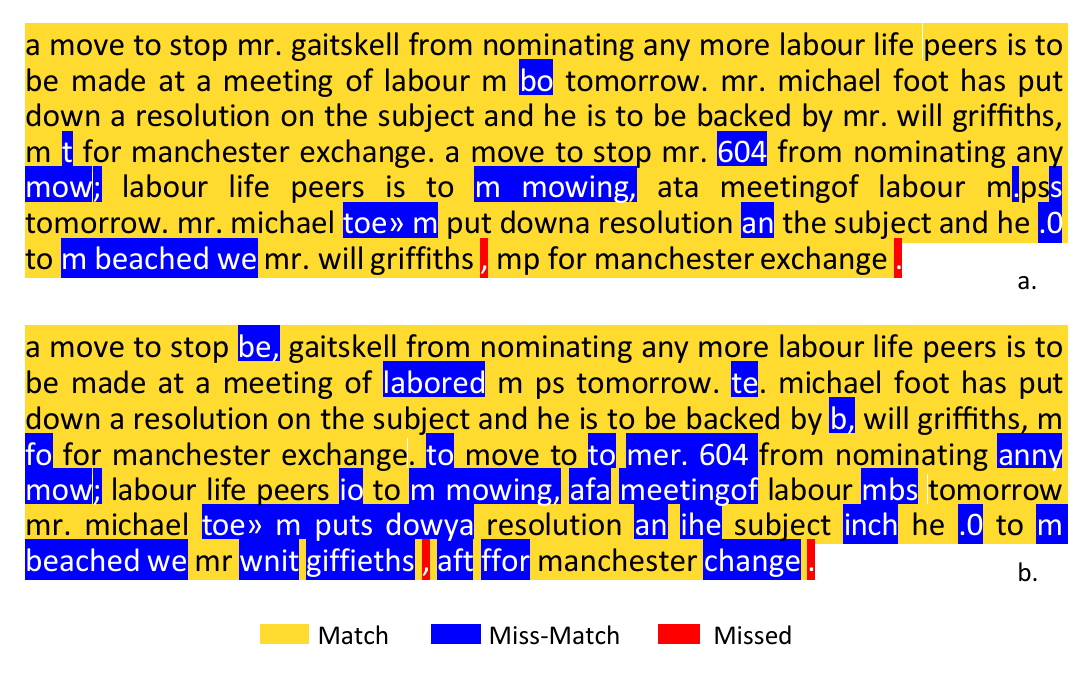}}
\vskip -0.1in
\caption{Effect of padding on text recognition: a. with padding, b. without padding.}
\label{fig:padding}
\end{center}
\vskip -0.3in
\end{figure} 
\begin{figure}
\begin{center}
\centerline{\includegraphics[width=9cm]{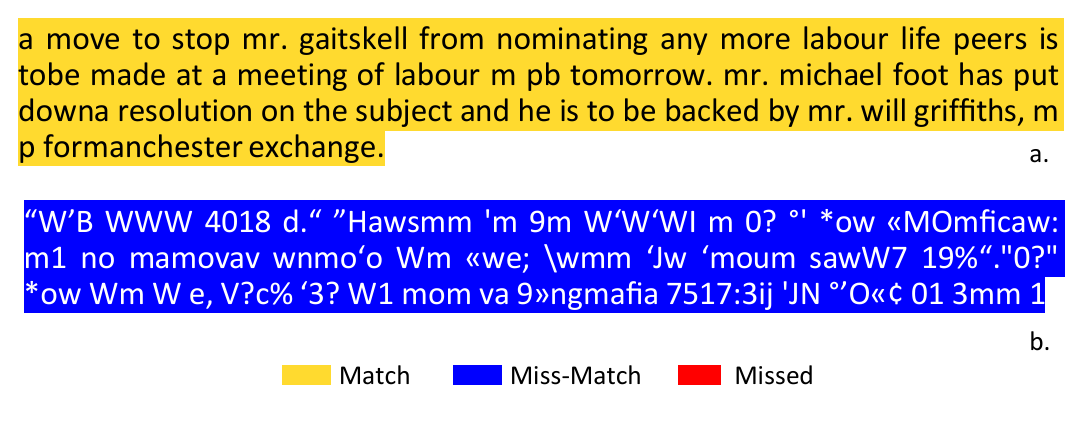}}
\vskip -0.1in
\caption{Effect of rotation on text recognition: a. with rotation, b. without rotation}
\label{fig:rotation}
\end{center}
\vskip -0.3in
\end{figure} 

\subsection{Analysis}
\noindent In this subsection, we investigate the impact of the preprocessing stages proposed in the process flow as discussed in the Section \ref{Methodology}.
The analysis is applied on a single page on the IAM database with the MIXED-IAM labels. 
Fig. \ref{fig:enhancement} shows the generated transcription with and without the image enhancement compared to the target transcription. 
The Levenshtein distance normalized by the length of the longer document between the prediction and the target achieved 89.27\% with enhancement. On the other hand, without enhancement the accuracy is drastically degraded to 54.79\%. 

Fig. \ref{fig:skew} shows the effect of the deskewness (vertically aligned) on a single image transcription.
As the figure illustrates, some parts of the text are missed in the OCR process if page is skewed. The truncated words degrade the overall accuracy of the transcription, where the recognition accuracy of this sample page with deskewness was 89.27\% and 85.44\% without it. 

Through our recognition process, words which fail the machine-printed recognition stage are cropped from the image and forwarded to the handwritten text recognition for further processing. During cropping, the text may fall at the image boundaries, which make it difficult for the handwriting text recognition to capture the pixels context and consequently the word boundaries accurately. To resolve this issue we pad the cropped images with white spaces in each direction. The effect of the padding is shown in Fig. \ref{fig:padding}, where the padded version achieved an accuracy of 89.27\% 
compared an accuracy of 82.57\% without the padding. 

Fig. \ref{fig:rotation} shows the transcription of the system for a vertically oriented page against a 180$^{\circ}$ rotation for the same page. It is clear from the figure that the transcription failed to generate a single correct word compared to the correctly aligned variant. 

\subsection{Results}
\noindent The evaluation of our proposed flow use the IAM database with the MIXED-IAM labels. We use several metrics to measure the performance of the system over different transcription levels, i.e. character, word and document.

For the character level evaluation, the Levenshtein distance between the predicted raw string and the target label is computed. The edit distance is then normalized using the longest length of the two strings. To evaluate the document level similarity, we extract the document embedding for both the target and the prediction, where the cosine distance is used to measure the similarity between the generated vectors. It is worth mentioning that the preprocessing stages execution time approach approaches 38 seconds on average per page, which is an optimization consideration for future work.

Table \ref{tab:chardoc} shows the performance evaluation of Levenshtein distance and document embeddings similarity for our proposed architecture using both the rule-based and context-based nomination. The table shows a transcription accuracy of 79.38\% using Levenshtein distance in combination with Rule-based nomination, with a comparable accuracy of 77.17\% for Context-based nomination. Higher accuracy could be achieved with Context-based nomination through considering other variants of pre-trained embedding models, used in the bi-gram embeddings generation, or a combination of them, which will be considered further in future work.
\begin{table}[!t]
\caption {Performance Evaluation using Character Level and Document Level Comparison}
\label{tab:chardoc}
\centering
 \begin{tabular}{ l c c}
       \toprule
 Evaluation Metric & Context-Based & Rule-Based\\
                   &  Accuracy(\%) &  Accuracy(\%) \\ 
       \midrule
 Levenshtein  &77.17 &79.38 \\ 
 Document Similarity & 75.06 &  69.77 \\  
        \bottomrule 
 
\end{tabular}
\end{table}

\begin{table}[!t]
\caption{Precision, Recall and F-Score of Proposed Architecture over IAM database}
\label{tab:bow}
\centering
 \begin{tabular}{ l c  c  c }
       \toprule
 Method & Precision(\%) & Recall(\%) & F-Score(\%)  \\ 
       \midrule
 Context-Based & 65.97 & 67.31& 66.46 \\  
 Rule-Based &  70.17& 68.14& 68.72  \\  
 
         \bottomrule 
 \end{tabular}
\end{table}

The BoW evaluation involved creating a set of distinct words using both the target and the predicted documents. The set acts as a dictionary, where each word in the vocabulary is assigned a unique identifier. Each of the two documents are further transformed using the established dictionary into a histogram composed of the unique identifiers present in the document and the frequency of their occurrences disregarding the syntax structure of the document. At this stage, the generated histograms for both the predicted and target documents can be compared against each other. Table \ref{tab:bow} shows the average precision, recall and F-score of our architecture for transcription of all documents of the IAM database using either the Rule-Based or the Context-Based nomination. The close values  of the precision and recall listed in Table \ref{tab:bow}, validates the stability of the reported accuracies of our proposed process flow.

\begin{table}[!t]
\caption{Number of Documents in Different Accuracy Ranges Using Context-Based Elimination Method.}
\label{tbl:heatmapcontext}
\centerline{\includegraphics[width=7cm]{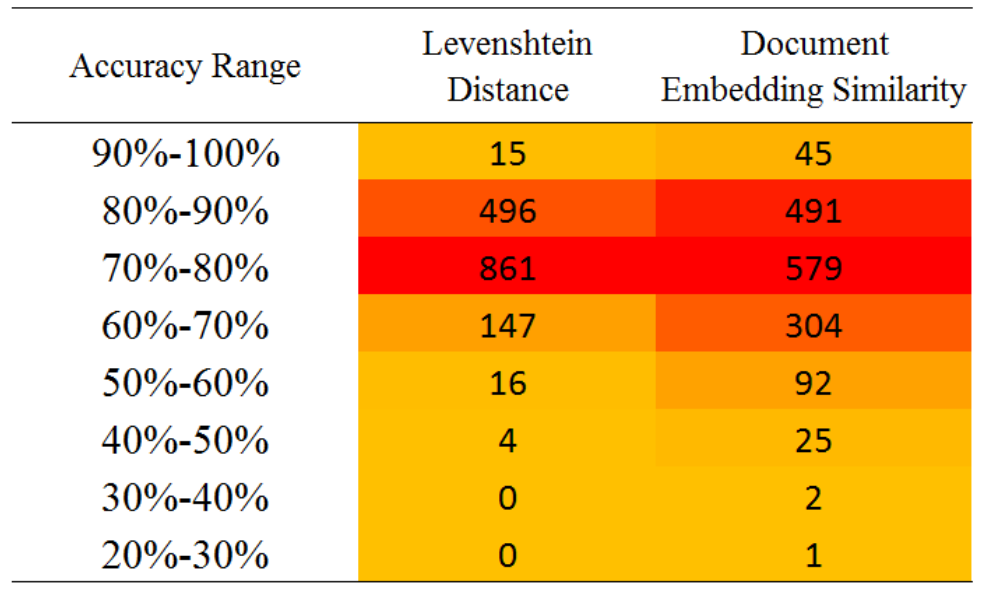}}
\vskip -0.01in
\end{table} 
\begin{table}[!t]
\caption{Number of Documents in Different Accuracy Ranges Using Rule-Based Elimination Method.}
\label{tbl:heatmaprule}
\centerline{\includegraphics[width=7cm]{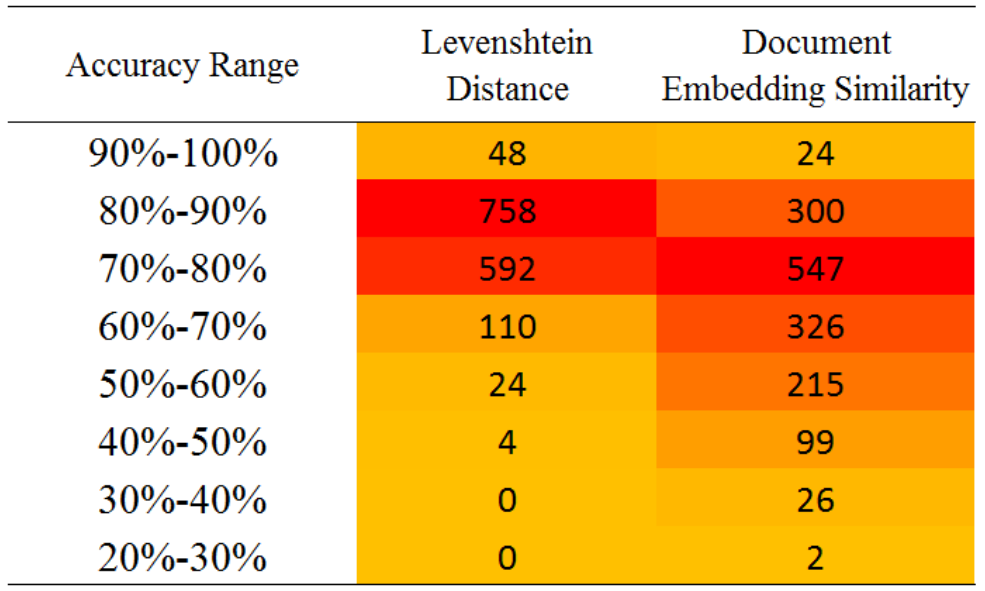}}
\vskip -0.01in
\end{table} 

Tables \ref{tbl:heatmapcontext} and \ref{tbl:heatmaprule} show the distribution of the 1539 documents of IAM database across different accuracy ranges using Context-Based and Rule-Based option nomination, respectively. Using the Levenshtein distance as a distance measure, 758 documents were transcribed with an average accuracy of 85\% compared to 496 documents using the Context-Based elimination. Despite the higher number of documents residing in the 80\%-90\% range for the Context-Based elimination compared to the Rule-Based using the embedding similarity as a distance measure, this is not indicative of better performance compared to the Levenshtein evaluation, but rather it is showing a variant evaluation representation that depends on the distance measure between the vector representation of documents. Future work will consider optimizing the Context-Based elimination to further exploit the context window to decide on the optimum candidate word.  

The recognition of each document through the process flow, results in creating option lists with one, three or four words for each word in a document based on the success or failure of machine-printed or handwritten recognition and the spell-checkers throughout the process.
Retrieving the optimum candidate word from the list generated for each word, occurs at the nomination stage, and has considerable impact on the resulting transcription. Following our analysis of the portion of words falling in each category of lists over the whole IAM database, we found that $65\%$ of the words are transcribed correctly through the machine-printed recognition (options list contains one word), $16\%$ of the words passed the handwritten recognition spell-checking (options list contains three words) and $19\%$ failed the spell-checking of hand-written recognized words (options list contain four words). These statistics are primarily dependent on the modules integrated in the process flow. Future work will consider more optimized recognizers and spell-checkers in addition to enhancing the context-based elimination. 

\section{Conclusions and Future Work}\label{Conclusions and Future Work}
\noindent We have proposed a process flow for Transcribing MIXed handwriting and machine-printed Text (TMIXT) in scanned text documents.
The TMIXT process flow allows recognizing text in an image of a document without the need for prior text categorization. 
The proposed process exploits several open-source libraries to investigate the feasibility of its implementation. However, the flow is generic enough to allow substituting any of the libraries used with different ones fulfilling the same task at the relevant stage,
which extends the ability of the TMIXT flow to be applied to other languages, or make use of more powerful tooling as they become available.  
We have evaluated the accuracy of TMIXT using different evaluation metrics based on character and word level in addition to document embeddings using the widely adapted IAM database.
Future work, will investigate enhancing the preprocessing stages for the input image and further optimization for the models used in text recognition either machine-printed or handwriting.
We will also consider adapting more enhanced methods for the spell-checking and different pre-trained models to generate the bi-gram embeddings used in the context-based nomination.

%%---------------------------------------------------------------------------- %
\section{Acknowledgement}
\noindent This work was funded by Applications Service Development  Operations Team,  Joint Forces Command - Information Systems and Services (ISS), UK.
%%---------------------------------------------------------------------------- %
\bibliographystyle{IEEEtran}
\balance
\bibliography{bib/Reference}

%%---------------------------------------------------------------------------- %
%\clearpage

\end{document}